%% file: preprint.tex
\documentclass[preprint, prX]{revtex4}

\usepackage{amsmath}    
\usepackage{graphicx}   
\usepackage{verbatim}   
\usepackage{color}      
\usepackage{subfigure}  
\usepackage{hyperref}   
\usepackage{amssymb}    
\usepackage{epsfig}
\usepackage{graphics,graphicx}
\usepackage{setspace}
\usepackage{url}
\usepackage{algorithm,algorithmic}

\begin{document}

\begin{abstract}
Memetic computation (MC) has emerged recently as a new paradigm of efficient algorithms for solving the hardest optimization problems. On the other hand, artificial bees colony (ABC) algorithms demonstrate good performances when solving continuous and combinatorial optimization problems. This study tries to use these technologies under the same roof. As a result, a memetic ABC (MABC) algorithm has been developed that is hybridized with two local search heuristics: the Nelder-Mead algorithm (NMA) and the random walk with direction exploitation (RWDE). The former is attended more towards exploration, while the latter more towards exploitation of the search space. The stochastic adaptation rule was employed in order to control the balancing between exploration and exploitation. This MABC algorithm was applied to a Special suite on Large Scale Continuous Global Optimization at the 2012 IEEE Congress on Evolutionary Computation. The obtained results the MABC are comparable with the results of DECC-G, DECC-G*, and MLCC.

\textit{To cite paper as follows: I. Fister, I. Fister Jr., J. Brest, V. Zumer, Memetic Artificial Bee Colony Algorithm for Large-Scale Global Optimization,
in Proc. IEEE Congress on Evolutionary Computation, Brisbane, Australia,  2012}

\end{abstract}

\title{Memetic Artificial Bee Colony Algorithm for\\Large-Scale Global Optimization}

\author{Iztok Fister}
\altaffiliation{University of Maribor, Faculty of electrical engineering and computer science
Smetanova 17, 2000 Maribor}
\email{iztok.fister@uni-mb.si}

\author{Iztok Fister Jr.}
\altaffiliation{University of Maribor, Faculty of electrical engineering and computer science
Smetanova 17, 2000 Maribor}
\email{iztok.fister@guest.arnes.si}

\author{Janez Brest}
\altaffiliation{University of Maribor, Faculty of electrical engineering and computer science
Smetanova 17, 2000 Maribor}
\email{janez.brest@uni-mb.si}

\author{Viljem \v{Z}umer}
\altaffiliation{University of Maribor, Faculty of electrical engineering and computer science
Smetanova 17, 2000 Maribor}
\email{zumer@uni-mb.si}

\maketitle

\section{Introduction}

A large-scale global optimization problem is defined as follows. Let us assume, an objective function $f(x)$ is given, where $x=(x_{1},\ldots,x_{D})$ is a vector of $D$ design variables in a decision space $S$. The design variables $x_{i}\in \{x_{lb},x_{ub}\}$ are limited by their lower $x_{lb}\in R$ and upper bounds $x_{ub} \in R$. The task of optimization is to find the minimum of the objective function. Note that large-scale refers to a huge number of variables, i.e. $n \geq 1,000$. Moreover, this problem belongs to the domain of function optimization.

The domain of function optimization serves as a 'test-bed' for many new comparisons and new features of various algorithms~\cite{Michalewicz:1996}. Over the past decade, different kinds of meta-heuristic optimization algorithms~\cite{Glover:2003,Michalewicz:2004} have been developed for solving this problem, for example: Simulated Annealing (SA)~\cite{Kirkpatrick:1983,Nolte:2000}, Evolutionary Algorithms (EAs)~\cite{Baeck:1996,Baeck:1997,Coello:2007,Lozano:2011}, Differential Evolution (DE)~\cite{Price:2005,Brest:2006,Brest:2011,Das:2011}, Particle Swarm Optimization (PSO)~\cite{Eberhart:1995,Clerc:2006} and Ant Colony Optimization (ACO)~\cite{Dorigo:1996,Dorigo:2004}. EAs hybridized with local search algorithms~\cite{Aarts:1997} have been successful within this domain. These kinds of EAs are often named as \textit{memetic algorithms} (MAs)~\cite{Moscato:1989,Merz:2000}. Memetic computation (MC)~\cite{Ong:2010} has emerged recently as a widening of MAs. MC uses a composition of \textit{memes} that represent those units of information encoded within \textit{complex structures} and interact with each other for the purpose of problem-solving. Namely, the meme denotes an abstract concept that can be, for example: a strategy, an operator or a local search algorithm.

As analog to EA, swarm intelligence (SI)~\cite{Blum:2008} has been inspired by nature. For foraging and defending, it simulates the collective behavior of social insects, flock of birds, ant-colonies, and fish schools. These particles are looking for good food sources and due to the interaction  between them move to the more promising regions within an environment. This concept has been exploited also by the Artificial Bees Colony (ABC) algorithm proposed by Karaboga and Basturk~\cite{Karaboga:2009} and achieved excellent results when solving continuous optimization problems~\cite{Karaboga:2007} as well as combinatorial optimization problems~\cite{Tasgetiren:2011,Fister:2012}.

Artificial bees form colonies within which each bee has limited capabilities and limited knowledge of its environment. However, such a colony is capable of developing a collective intelligence that is used for foraging. The foraging process is divided into three components: food sources, employed foragers, and unemployed foragers. On the other hand, Iacca et al.~\cite{Iacca:2011} assert that an optimal memetic exploration of the search space is composed of three memes: the first stochastic with long-search radius, the second stochastic with a moderate-search radius, and the third deterministic with short-search radius. In line with this, it has been identified that the original ABC algorithm also explores the search space over three stages that could correspond to three memes: the first stochastic with a long-search radius (food sources), the second stochastic with a moderate-search radius (employed foragers), and the third random with long-search radius (unemployed foragers).

Unfortunately, the deterministic meme with a short-search radius is missing from this identification. Therefore, the original ABC algorithm has been hybridized with the local search heuristics representing this deterministic meme. Two local search methods were applied within the proposed memetic ABC (MABC) algorithm: the Nelder/Mead (NM) simplex method~\cite{Rao:2009} and the random walk method with direction exploitation (RWDE)~\cite{Rao:2009}. The former is designed more towards exploration, whilst the latter more towards exploitation of the continuous search space. The stochastic adaptive rule as specified by Neri~\cite{Neri:2011} was applied for balancing the exploration and exploitation.

In order to prevent a slow convergence of the original ABC algorithm caused because of a single-dimension update by the crossover operator, a new crossover operator capable of updating multiple dimensions was developed within MABC. Additionally, new mutation strategies similar to DE strategies were incorporated into this algorithm.

Finally, this MABC algorithm was applied to a Special suite on Large Scale Continuous Global Optimization at the 2012 IEEE Congress on Evolutionary Computation.

This paper is structured as follows. Section II describes the MABC algorithm. In Section III, the test suite together with the experimental setup is discussed in detail. In Section IV, the obtained results are presented and compared with other algorithms. In Section V, the final remarks are presented, and a look is taken on future work.

\section{Memetic Artificial Bees Colony Algorithm for Large-Scale Global Optimization}

In the original ABC algorithm, the foraging process is performed by three groups of bees: employed bees, onlookers, and scouts. Each food source is discovered by only one employed bee. On the basis of information obtained by the employed foragers, the onlooker bees make a decision, which food source to visit. When the food source is exhausted, the corresponding unemployed foragers become scouts.

In order to make a super-fit individual~\cite{Neri:2011} that guides the direction in which the search space needs to be explored by the bee colony, an additional step is added to the original ABC algorithm, i.e. locally improving the best bee. As a result, the final scheme for the MABC algorithm is obtained as follows:

\begin{algorithm}[htb]
\caption{Pseudo code of the MABC algorithm}
\label{alg:prog}
\begin{algorithmic}[1]
\STATE Init();
\WHILE {!TerminationConditionMeet()}
\STATE SendEmployedBees();
\STATE SendOnlookerBees();
\STATE LocalImproveBestBee();
\STATE SendScouts();
\ENDWHILE
\end{algorithmic}
\end{algorithm}

MABC is a population-based algorithm of size $\mathit{NP}$, where the candidate solutions $x_{i}$ are vectors of $D$ design variables within a decision space $S$. Initial population (function Init() in Algorithm~\ref{alg:prog}) is randomly sampled within decision space $S$, as follows:
\begin{equation}
\label{eq:init}
    x^{(0)}_{i,j} = \begin{matrix}
    \mathrm{rand} \left ( 0,1 \right ) . \left ( ub-lb \right ) + lb, & \mathrm{for}\ j=1\ldots D,
\end{matrix}
\end{equation}

\noindent where function $\mathit{rand} \left ( 0,1 \right )$ returns a random value between 0 and 1, and \textit{lb}, \textit{ub} denote the lower and upper bounds of the candidate solution $x_{i}$.

The core of the foraging process (sequence of function calls in the \textbf{while} loop in Algorithm~\ref{alg:prog}) consists of four functions:
\begin{itemize}
  \item SendEmployedBees(): sending the employed bees onto the food sources and evaluating their nectar amounts,
  \item SendOnlookerBees(): sharing information about food sources with employed bees, selecting the proper food source and evaluating their nectar amounts,
  \item LocalImproveBestBee(): determining the best food source and improve it,
  \item SendScouts(): determining the scout bees and then sending them to find new possibly better food sources.
\end{itemize}

In the sense of MC analysis due to~\cite{Iacca:2011}, each of these four functions perform certain exploration task, as follows:
\begin{itemize}
  \item stochastic long-distance exploration (function SendEmployedBees()),
  \item stochastic moderate-distance exploration (function SendOnlookerBees()),
  \item deterministic short-distance exploration (function LocalImproveBestBee()),
  \item random long-distance exploration (function SendScouts()).
\end{itemize}

In continuation, these exploration tasks are discussed in detail.

\subsection{Stochastic long-distance exploration}
The stochastic long-distance exploration is performed in the DE fashion~\cite{Price:2005} as follows. In order to obtain a trial solution, three operators are applied on each candidate's solution, i.e. mutation, crossover and selection. The mutation operator in MABC implements a $'\mathit{rand}/1/\mathit{bin}'$ like DE mutation strategy, according to Eq.~(\ref{eq:strat1}):
\begin{equation}\label{eq:strat1}
    v_{i}^{(t)}=x_{r1}^{(t)}+\left ( x_{r2}^{(t)}-x_{r3}^{(t)} \right ),
\end{equation}

\noindent where $r1$, $r2$, $r3$ denote randomly-selected candidate solutions, and $t$ is the generation number. This operator is slightly different from the original ABC modification operator. Conversely to DE, the constant scale parameter $F$ is unused in the Eq.~(\ref{eq:strat1}).

In place of the original ABC expression for producing a new food position that modifies only single dimension of trial solution, the new expression is developed in MABC that is capable to modify multiple dimensions in trial solution. The number of modifications are controlled by $\mathit{CR}$ parameter as follows:
\begin{equation}\label{eq:xover1}
    u_{i,j}^{(t)}=\left\{\begin{matrix}
    v_{i,j}^{(t)} & \mathrm{if} \left ( \mathrm{rand}_j(0,1)\leq \mathit{CR}\ ||\ j==j_{r4} \right ), \\
    x_{i,j}^{(t)} & \mathrm{otherwise},
    \end{matrix}\right.
\end{equation}

\noindent where $j_{r4}$ denotes a random dimension within the trial solution. According to the minimum value of the objective function, the selection operator selects the best between the candidate and trial solution, as expressed in Eq.~(\ref{eq:sel1}):
\begin{equation}\label{eq:sel1}
    x_{i}^{t+1}=\left\{\begin{matrix}
    u_{i}^{(t)} & \mathrm{if} \left ( f\left ( u_{i}^{(t)} \right ) \leq f\left ( x_{i}^{(t)} \right )\right ) \\
    x_{i,j}^{(t)} & \mathrm{otherwise}.
\end{matrix}\right.
\end{equation}

The long-distance exploration comprises two stochastic mechanisms: the mutation strategy and crossover operators. The first mechanism focuses on exploration because the modified dimensions of the trial solution is independent of the current value, i.e. the new value is sampled as three dimensions of randomly-selected candidate solutions. The second mechanism determines how many modifications should be applied to the trial solution. As a result, this exploration attempts to detect new promising regions within the entire decision space $S$.

\subsection{Stochastic moderate-distance exploration}
Onlooker bees visit a food source after making a decision if the appropriate food source lacks sufficient nectar amounts. Within more nectar amounts, more bees can be expected in the vicinity of this food source. The property $q$ that the $i$-th onlooker bee should visit the nectar amount is expressed as follows:
\begin{equation}
\label{eq:prop}
    q_{i}=\frac{f_{i}-f_{worst}}{f_{best}- f_{worst}},
\end{equation}

\noindent where $f_{i}$, $f_{worst}$ and $f_{best}$ denotes the nectar amounts of $i$-th, $worst$ and $best$ food sources.

The stochastic moderate-distance exploration is performed similarly to long-distance exploration, i.e. a trial solution undergoes acting the three operators: mutation, crossover and selection. Here, MABC implements a $'\mathit{currenttobest}/1/\mathit{bin}'$ like DE strategy as the mutation operator, expressed as:
\begin{equation}
\scriptsize
\label{eq:strat2}
    v_{i}^{(t)}=
\begin{matrix}
    x_{i}^{(t)}+\left ( x_{best}^{(t)}-x_{i}^{(t)} \right )+\left ( x_{r1}^{(t)}-x_{r2}^{(t)} \right ) & \begin{cases}
 & \mathrm{if} \left ( q_{i} < r0 \right ) \\
 & \mathrm{else}\ i=i+1
\end{cases}
\end{matrix}
\normalsize
\end{equation}

\noindent where $r0$ denotes the randomly generated number between 0 and 1 ($\mathrm{rand}(0,1)$), $r1$ and $r2$ are the randomly-selected candidate solutions and $q_{i}$ is expressed according to Eq.~(\ref{eq:prop}). The crossover and selection operators are implemented according to Eqs~(\ref{eq:xover1}) and~(\ref{eq:sel1}).

Note that the trial solution is only generated if the nectar amounts are sufficient. If not, the next food source is selected. This process repeats until the number of onlooker bees does not exceed $\mathit{NP}$. In this manner, the food sources with more nectar amounts are likely to be visited by onlookers. As a results, onlookers are collected in the vicinity of more promising food sources.

In summary, the function SendOnlookerBees() directs onlookers to more promising regions of the search space. The applied mutation operator also contributes to this directed search.

\subsection{Deterministic short-distance exploration}

The deterministic short-distance exploration tries to fully exploit promising search directions. The goal of this exploration is to bring a candidate solution into the local optimum. In the case that the solution is the global optimum, the search needs to be finished. In MABC, short-distance exploration concerns for maintaining a \textit{diversity} of population.

In population-based algorithms, diversity plays a crucial role in the success of the optimization~\cite{Neri:2011a}. The diversity of a population is a measure of the \textit{different} solutions present~\cite{Eiben:2003}. This measure can be expressed as the number of different fitness values present, the number of different phenotypes present, or the number of different genotypes present. In this study, the fitness diversity metric is expressed as~\cite{Neri:2O07}:
\begin{equation}
\label{eq:metric}
    \mathit{\Psi} = 1- \left | \frac{f_{avg}-f_{best}}{f_{worst}-f_{best}} \right |,
\end{equation}

\noindent where $f_{avg}$, $f_{best}$, and $f_{worst}$ denote the average, best, and worst fitness of individuals in the population. Note that the metric returns values between 0 and 1.

The main characteristic of this metric is that its value is independent on the range of variability the fitness values. It is very sensible to small variations and thus, especially suitable for fitness landscapes containing \textit{plateaus} and \textit{low gradient} areas~\cite{Neri:2011a}. Although more fitness diversity metrics were used in the experiments, this metric demonstrated that it is the most suitable for the MABC algorithm.

In the sense of deterministic short-distance exploration, two local search algorithms were developed that run within the ABC framework. The first is the Nelder-Mead Algorithm (NMA)~\cite{Nelder:1965} with exploration features and the second the Random Walk with Direction Exploitation (RWDE)~\cite{Rao:2009} with exploitation features. In order to coordinate the exploration/exploitation process, the following exponential distribution is used:
\begin{equation}
\label{eq:dist}
    p \left ( \mathit{\Psi} \right ) = e^{\frac{-\left ( \mathit{\Psi} - \mu_{p} \right )}{2\sigma^{2}_{p}}},
\end{equation}

\noindent where $\mathit{\Psi}$ are the fitness diversity metric, and variable $\mu_{p}$ denotes the mean value and $\sigma_{p}$ is the standard deviation of diversity metric $\mathit{\Psi}$ calculated so far. Both values $\mu_{p}$ and $\sigma_{p}$ are calculated incrementally in each generation according to the Knuth's algorithm~\cite{Knuth:1981}.

The following adaptive scheme is applied for balancing between both local search algorithms:
\begin{equation}\label{eq:sch1}
    \mathrm{if}\left\{\begin{matrix}
    \mathrm{rand} \left ( 0,1 \right ) > p \left ( \mathit{\Psi} \right ) & => & \textrm{use NMA} \\
    \mathrm{otherwise} & => & \textrm{use RWDE},
\end{matrix}\right.
\end{equation}

\noindent where $\mathrm{rand} \left ( 0,1 \right )$ denotes a randomly generated value between 0 and 1. From Eq.~(\ref{eq:sch1}) it can be seen that when the population looses the diversity, i.e. $p \left ( \mathit{\Psi} \right ) > 0.5$, the NMA exploration algorithm is executed, whilst in contrary the RWDE exploitation algorithm takes the initiative.

\subsection{Random long-distance exploration}
In nature, when the food source is exhausted the foragers become scouts. These scouts look for a new food sources within the environment. In MABC, scouts are simulated by new randomly generated food source that should be discovered by the foragers in the remaining cycles of this algorithm. The food is exhausted when the predefined number of cycles (also named scout \textit{limit}) has expired, in which no further improvement of the nectar amounts are detected.

The scouts are generated according to Eq.~(\ref{eq:init}). Note that this exploration is treated as long-distance because the new food sources are sampled within the whole decision space $S$. On the other hand, the exploration is blind without any history information explored so far.

\subsection{MABC framework}
In the spirit of MC, the mentioned kinds of explorations can be seen as memes. These memes interact between each other in order to problem-solving. In MABC, the original ABC framework does not exploit the MC paradigm as a whole. That is, the memes are executed sequential within each generation. Let us emphasis that the stochastic long-distance (function SendEmployesBees()) and the stochastic moderate-distance (function SendOnlookerBees()) explorations are performed unconditionally, i.e. in each generation. On the other hand, the performances of the deterministic short-distance (function LocalImproveBestBee()) and the random long-distance (function SendScouts()) explorations are controlled by the parameters local search $ratio$ and scouts $limit$. The first parameter regulates the ratio between the global and local search, whilst the second determines when to start exploring new regions of the continuous search space.

\section{Experiments}
A goal of the experiments was the applying of MABC to the specific test-suite proposed in the Special session on Large Scale Continuous Global Optimization at the 2012 IEEE Congress on Evolutionary Computation, the obtaining of results, and then comparing these with the results of DECC-G*, DECC-G, and MLCC as proposed by the organizing committee, to see how many competitive results were achieved.

\subsection{Test suite}

The test-suite consists of five problem classes, i.e. various high-dimensional problems ($\mathit{D}=1,000$):

\small
\begin{enumerate}
  \item Separable functions:
  \begin{itemize}
    \item $F_{1}$: Shifted Elliptic Function.
    \item $F_{2}$: Shifted Rastrigin's Function.
    \item $F_{3}$: Shifted Ackley's Function.
  \end{itemize}
  \item Single-group $m$-nonseparable functions ($m=50$):
  \begin{itemize}
    \item $F_{4}$: Single-group Shifted and $m$-rotated Elliptic Function.
    \item $F_{5}$: Single-group Shifted and $m$-rotated Rastrigin's Function.
    \item $F_{6}$: Single-group Shifted and $m$-rotated Ackley's Function.
    \item $F_{7}$: Single-group Shifted and $m$-dimensional Schwefel's Problem 1.2.
    \item $F_{8}$: Single-group Shifted and $m$-dimensional Rosenbrock's Function.
  \end{itemize}
  \item $\frac{D}{2m}$-group $m$-nonseparable functions ($m=50$):
  \begin{itemize}
    \item $F_{9}$: $\frac{D}{2m}$-group Shifted and $m$-rotated Elliptic Function.
    \item $F_{10}$: $\frac{D}{2m}$-group Shifted and $m$-rotated Rastrigin's Function.
    \item $F_{11}$: $\frac{D}{2m}$-group Shifted and $m$-rotated Ackley's Function.
    \item $F_{12}$: $\frac{D}{2m}$-group Shifted and $m$-dimensional Schwefel's Problem 1.2.
    \item $F_{13}$: $\frac{D}{2m}$-group Shifted and $m$-dimensional Rosenbrock's Function.
  \end{itemize}
  \item $\frac{D}{m}$-group $m$-nonseparable functions ($m=50$):
  \begin{itemize}
    \item $F_{14}$: $\frac{D}{m}$-group Shifted and $m$-rotated Elliptic Function.
    \item $F_{15}$: $\frac{D}{m}$-group Shifted and $m$-rotated Rastrigin's Function.
    \item $F_{16}$: $\frac{D}{m}$-group Shifted and $m$-rotated Ackley's Function.
    \item $F_{17}$: $\frac{D}{m}$-group Shifted and $m$-dimensional Schwefel's Problem 1.2.
    \item $F_{18}$: $\frac{D}{m}$-group Shifted and $m$-dimensional Rosenbrock's Function.
  \end{itemize}
  \item Fully-nonseparable:
  \begin{itemize}
    \item $F_{19}$: Shifted Schwefel's Problem 1.2.
    \item $F_{20}$: Shifted Rosenbrock's Function.
  \end{itemize}
\end{enumerate}
\normalsize

Note that the separability of a function determines how difficult the function is to solve. That is, the function $f(x)$ is separable if its parameters $x_{i}$ are independent. In general, the separable functions are considered to be the easiest. In contrast, the fully-nonseparable functions are usually the more difficult to solve~\cite{Tang:2009}. The degree of separability could be controlled by randomly dividing the object variables into several groups, each of which contains a particular number of variables. Although some of used functions are separable in their original forms, applying techniques as Salomon's random coordinate rotation make them non-separable. Furthermore, the global optimum of the function could also be shifted.

\subsection{Experimental setup}

Table~\ref{tab:tab_1} presents the characteristics of MABC used in the experiments. Note that all values of algorithm parameters presented in this table were the best found during extensive experiments.

\begin{table}[htb]        
\caption{Parameters used for MABC}
\label{tab:tab_1}
\vspace{-5mm}
\begin{center}
\begin{tabular}{ l l l l }
\hline
 Parameter & & & Value \\
\hline
Maximum number of runs & & & 25 \\
Maximum FEs & & & 3,000,000 \\
Population size & & & 20 \\
Crossover Probability $CR$ & & & 0.01 \\
Local search $ratio$ & & & 0.006 \\
Scouts $limit$ & & & 200 \\
\hline
\end{tabular}
\end{center}
\vspace{-5mm}
\end{table}

The number of independent runs of the MABC algorithm was limited to 25. The maximum number of function evaluations $\mathit{FEs}$ was fixed at 3 millions. Besides the final result, the error values by $\mathit{FEs}=120,000$ and $\mathit{FEs}=600,000$ were recorded. All these values were prescribed by the organizing committee of this competition.

The population size was set to 20. MABC, with higher population sizes, converged slowly, whilst the same algorithm with smaller population sizes remained in local optimum. The crossover probability $\mathit{CR}$ was set to 0.01. That is, each crossover modified 10 dimensions of the trial solution. The higher values of this parameter decreases the results of MABC drastically, whilst decreasing the number had a harmful impact on the results. The higher values for the local search ratio had deteriorating effect on the results. In this context, the value $r=0.006$ was demonstrated the best results. The scout's limit of 200 presented the best bias between the search progress (exploitation) and exploration of new regions within the search space.

\subsection{PC configuration}
All runs were made on HP Compaq with the following configurations:
\begin{enumerate}
  \item Processor - Intel Core i7-2600 3.4 (3.8) GHz
  \item RAM - 4GB DDR3
  \item Operating system - Linux Mint 12
\end{enumerate}

MABC was implemented within the Eclipse Indigo CDT framework.

\section{Results}
The purpose of this section is to illustrate the results obtained by MABC. The results are analyzed from different points of view and summarized within the following subsections:

\begin{itemize}
  \item convergence plots,
  \item summary of results and
  \item comparison with other algorithms.
\end{itemize}

The obtained results are presented in detail in the reminder of this paper.

\subsection{Convergence Plots}

When a convergence of the results by solving different functions is analyzed four characteristic curves can be obtained. The characteristic plots, for example, of the functions $F2$, $F11$, $F15$, and $F16$, are illustrated in Figs.~1-4. Note that the average results over 25 runs are considered in all plots.

\begin{figure}[htb]	
\centering
\label{fig:conv_1}
\caption{Convergence of mean value for function $F_{2}$}
\includegraphics[width=0.33\textwidth]{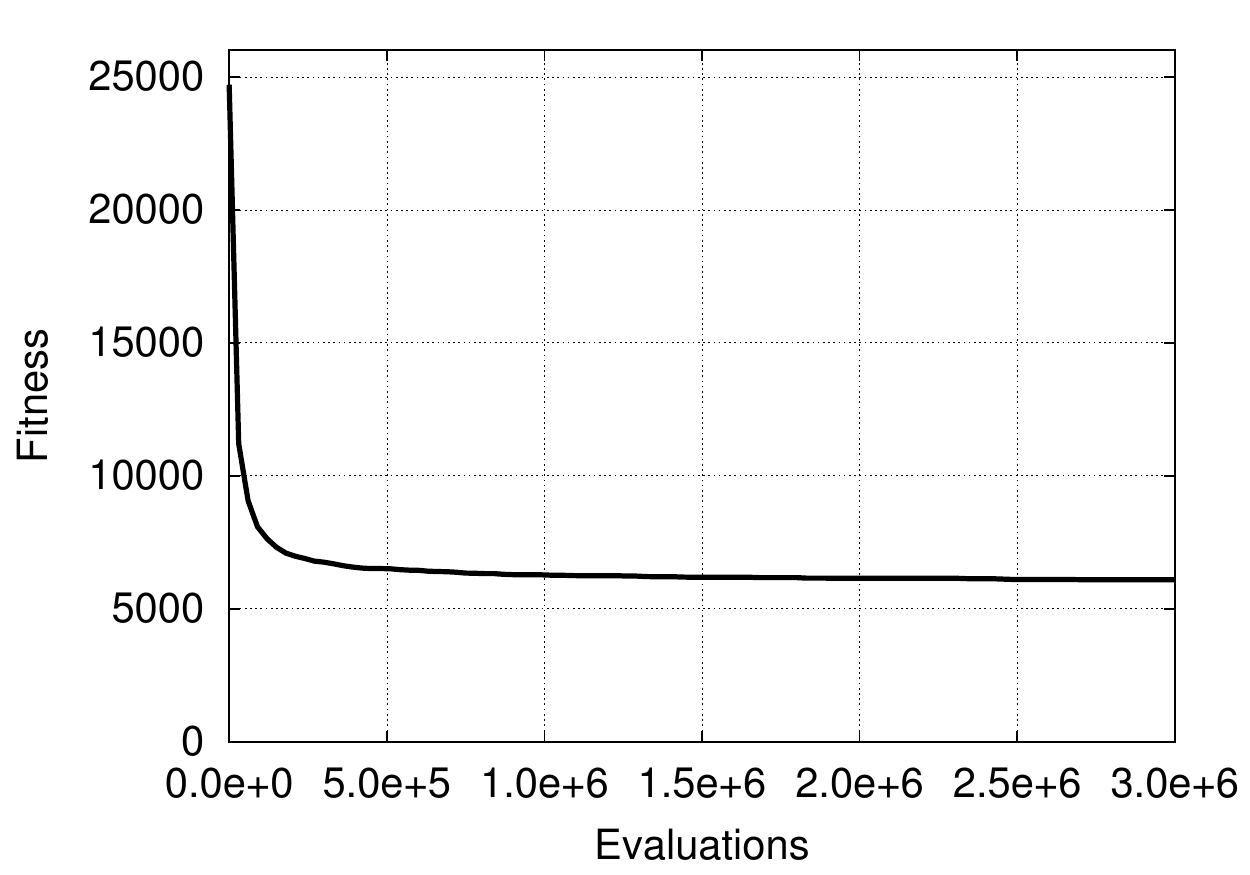}
\vspace{-5mm}
\end{figure}

\begin{figure}[htb]
\centering
\label{fig:conv_2}
\caption{Convergence of mean value for function $F_{11}$}
\includegraphics[width=0.33\textwidth]{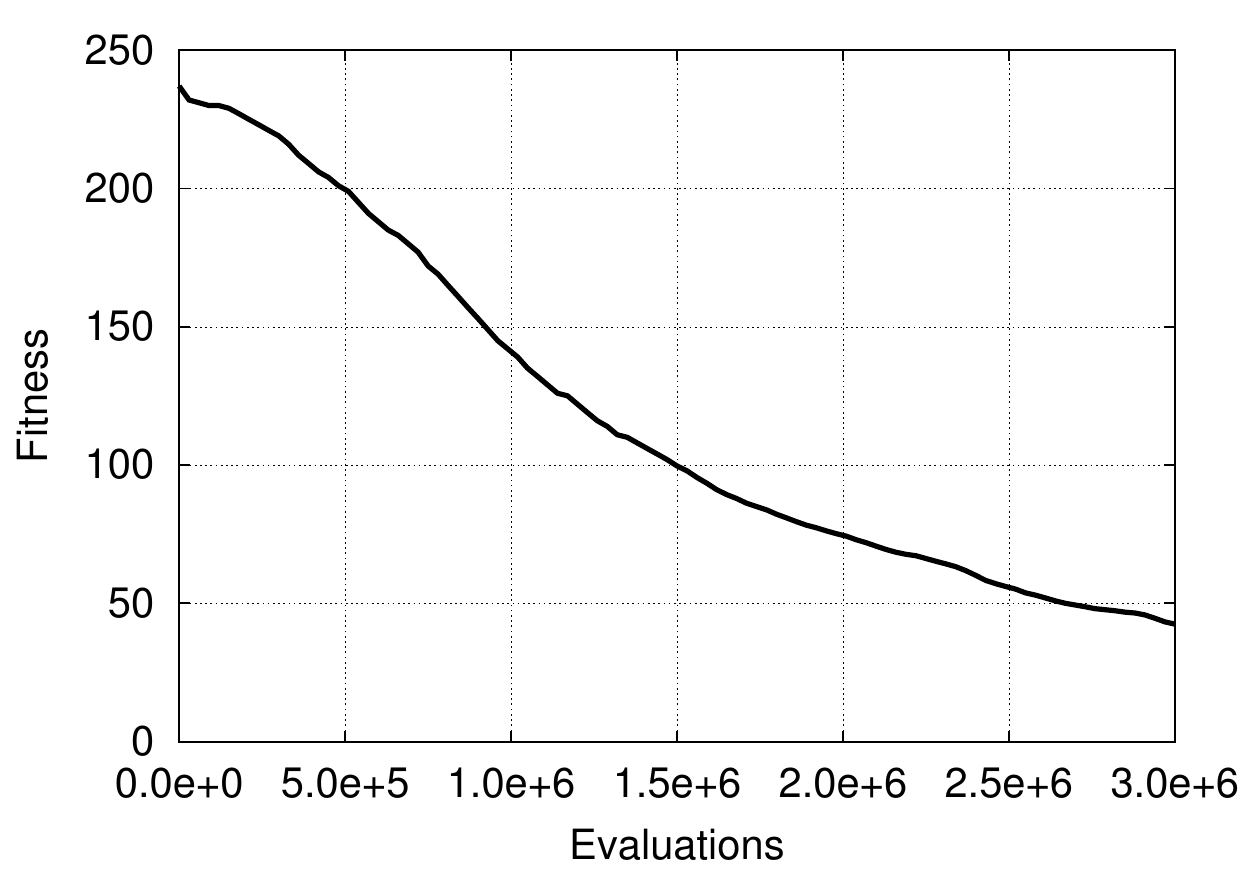}
\vspace{-5mm}
\end{figure}

\begin{figure}[htb]
\centering
\label{fig:conv_3}
\caption{Convergence of mean value for function $F_{15}$}
\includegraphics[width=0.33\textwidth]{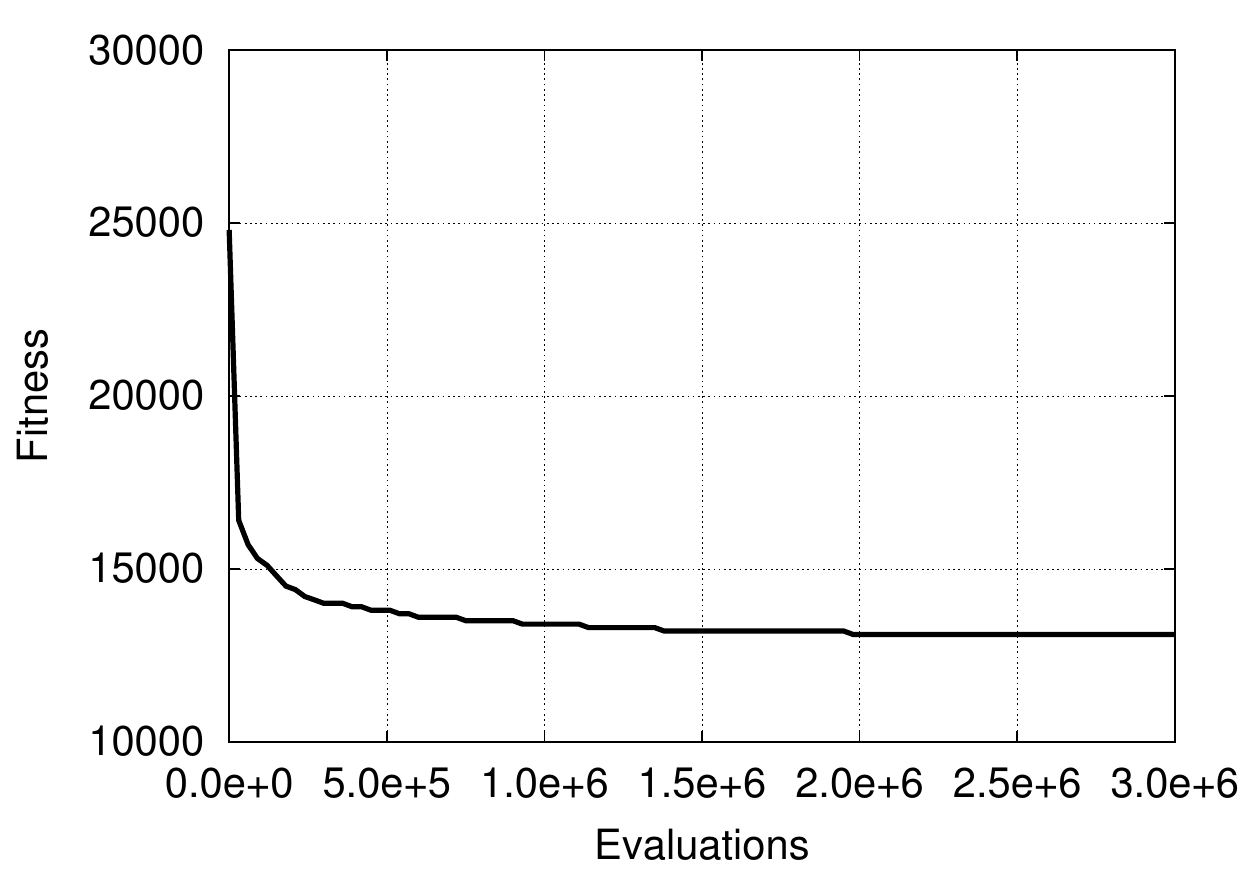}
\vspace{-5mm}
\end{figure}

\begin{figure}[htb]
\centering
\label{fig:conv_4}
\caption{Convergence of mean value for function $F_{16}$}
\includegraphics[width=0.33\textwidth]{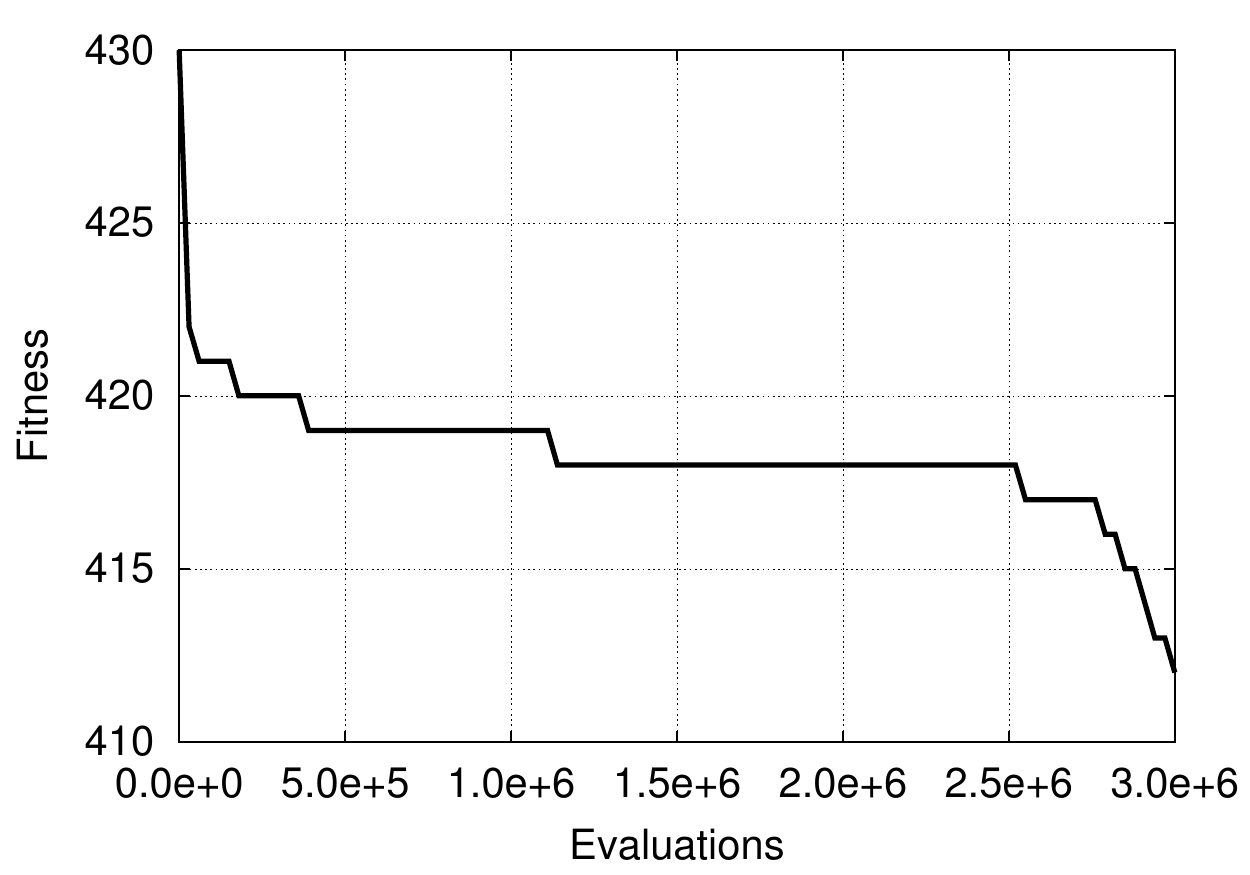}
\vspace{-5mm}
\end{figure}

The following characteristics can be observed on the basis of these convergence plots:
\begin{itemize}
  \item $F_{2}$: the convergence of this function is very fast until it becomes stuck within the local optimum.
  \item $F_{11}$: the convergence if this function is slow but continuous. It seems that the optimal value can be reached by increasing the maximum number of function evaluations.
  \item $F_{15}$: the convergence of this function is very fast during the first 500,000 evaluations, then, it converges slowly and  in the last third, of the evaluation numbers, stagnation is detected.
  \item $F_{16}$: this function converges in steps but its convergence is very slowly.
\end{itemize}

\subsection{Summary of Results}
The results of optimizing the test suite are illustrated in Table~\ref{tab:tab_2}, from which the following conclusions can be derived: \begin{itemize}
  \item Separable functions $F_{1}-F_{3}$ are the easiest to solve for MABC;
  \item Partial-separable Elliptic functions ($F_{4}, F_{9}, F_{14}$) are the hardest nut to crack for MABC;
  \item On average, the $\frac{D}{2m}$-group shifted and $m$-rotated functions are easier to solve than the $\frac{D}{m}$-group shifted and $m$-rotated, but the single-group shifted and $m$-rotated functions are the hardest to solve.
\end{itemize}

Even though, it should be valid that fully-nonseparable functions are the hardest to solve, this truth could not be seen from the results of the experiments.

\subsection{Comparison with other algorithms}
In this experiment, the results of the following algorithms are compared with the results of MABC (Table~\ref{tab:comp2}):
\begin{itemize}
  \item DECC-G,
  \item DECC-G* and
  \item MLCC.
\end{itemize}

\begin{table*}[!htb]
\begin{small}
\begin{center}
\caption{CEC-2012 Large Scale Continuous Global Optimization Problems}
\label{tab:tab_2}
\scalebox{0.7}
{
\begin{tabular}{ | c | c | c | c | c | c | c | c | c | }
\hline
 &  & $F_{1}$ & $F_{2}$ & $F_{3}$ & $F_{4}$ & $F_{5}$ & $F_{6}$ & $F_{7}$ \\
\hline
 & Best & 8.82e+06 & 7.02e+03 & 4.17e+00 & 1.31e+13 & 2.51e+08 & 2.34e+04 & 9.06e+09 \\
 & Median & 1.46e+07 & 7.60e+03 & 5.11e+00 & 3.00e+13 & 3.00e+08 & 1.46e+06 & 2.42e+10 \\
FEs = 1.2e5 & Worst & 3.07e+07 & 8.19e+03 & 6.31e+00 & 5.79e+13 & 4.06e+08 & 7.79e+06 & 3.61e+10 \\
 & Mean & 1.65e+07 & 7.63e+03 & 5.16e+00 & 3.20e+13 & 3.02e+08 & 1.63e+06 & 2.32e+10 \\
 & Std & 6.20e+06 & 2.48e+02 & 5.05e-01 & 1.30e+13 & 3.41e+07 & 1.60e+06 & 5.82e+09 \\
\hline
 & Best & 8.64e-04 & 5.93e+03 & 3.88e-05 & 1.49e+12 & 1.36e+08 & 3.61e-02 & 7.59e+07 \\
 & Median & 1.43e-02 & 6.45e+03 & 1.21e+00 & 3.34e+12 & 1.86e+08 & 2.52e+00 & 1.92e+08 \\
FEs = 6.0e5 & Worst & 3.52e+00 & 6.85e+03 & 2.84e+00 & 7.28e+12 & 2.32e+08 & 3.85e+00 & 1.87e+09 \\
 & Mean & 1.77e-01 & 6.44e+03 & 1.22e+00 & 3.82e+12 & 1.85e+08 & 2.34e+00 & 2.70e+08 \\
 & Std & 6.99e-01 & 2.29e+02 & 6.36e-01 & 1.81e+12 & 2.66e+07 & 9.69e-01 & 3.46e+08 \\
\hline
 & Best & 4.07e-22 & 5.82e+03 & 1.52e-12 & 4.52e+11 & 8.65e+07 & 7.61e-09 & 2.37e+03 \\
 & Median & 8.86e-22 & 6.08e+03 & 8.95e-01 & 1.44e+12 & 1.09e+08 & 1.98e+00 & 1.04e+04 \\
FEs = 3.0e6 & Worst & 5.44e-21 & 6.35e+03 & 2.79e+00 & 3.85e+12 & 1.56e+08 & 3.27e+00 & 9.85e+04 \\
 & Mean & 1.63e-21 & 6.09e+03 & 8.20e-01 & 1.64e+12 & 1.13e+08 & 1.86e+00 & 1.36e+04 \\
 & Std & 1.42e-21 & 1.29e+02 & 7.96e-01 & 9.92e+11 & 1.73e+07 & 8.49e-01 & 1.84e+04 \\
\hline
 &  & $F_{8}$ & $F_{9}$ & $F_{10}$ & $F_{11}$ & $F_{12}$ & $F_{13}$ & $F_{14}$ \\
\hline
 & Best & 2.72e+07 & 1.27e+09 & 1.21e+04 & 2.24e+02 & 1.07e+06 & 4.87e+05 & 3.28e+09 \\
 & Median & 1.01e+08 & 1.66e+09 & 1.29e+04 & 2.31e+02 & 1.33e+06 & 6.25e+05 & 3.82e+09 \\
FEs = 1.2e5 & Worst & 2.01e+08 & 1.98e+09 & 1.37e+04 & 2.32e+02 & 1.49e+06 & 3.21e+06 & 4.65e+09 \\
 & Mean & 1.17e+08 & 1.69e+09 & 1.29e+04 & 2.30e+02 & 1.32e+06 & 7.50e+05 & 3.93e+09 \\
 & Std & 5.09e+07 & 1.72e+08 & 3.97e+02 & 1.96e+00 & 1.09e+05 & 5.24e+05 & 3.73e+08 \\
\hline
 & Best & 3.44e+05 & 1.83e+08 & 1.02e+04 & 9.83e+01 & 2.09e+05 & 1.10e+03 & 5.83e+08 \\
 & Median & 7.32e+07 & 2.26e+08 & 1.10e+04 & 2.28e+02 & 2.59e+05 & 1.71e+03 & 6.68e+08 \\
FEs = 6.0e5 & Worst & 1.55e+08 & 2.68e+08 & 1.13e+04 & 2.30e+02 & 3.15e+05 & 3.72e+03 & 7.52e+08 \\
 & Mean & 6.27e+07 & 2.26e+08 & 1.09e+04 & 1.88e+02 & 2.54e+05 & 1.86e+03 & 6.64e+08 \\
 & Std & 4.39e+07 & 2.01e+07 & 2.42e+02 & 5.16e+01 & 2.21e+04 & 6.83e+02 & 4.32e+07 \\
\hline
 & Best & 1.14e+03 & 3.06e+07 & 1.02e+04 & 1.35e+01 & 1.01e+04 & 5.34e+02 & 1.04e+08 \\
 & Median & 5.91e+05 & 3.74e+07 & 1.05e+04 & 3.17e+01 & 1.27e+04 & 7.99e+02 & 1.23e+08 \\
FEs = 3.0e6 & Worst & 7.87e+07 & 4.39e+07 & 1.08e+04 & 1.80e+02 & 1.54e+04 & 1.41e+03 & 1.31e+08 \\
 & Mean & 9.77e+06 & 3.76e+07 & 1.04e+04 & 4.25e+01 & 1.28e+04 & 8.53e+02 & 1.22e+08 \\
 & Std & 2.12e+07 & 3.43e+06 & 1.84e+02 & 3.40e+01 & 1.36e+03 & 2.19e+02 & 6.78e+06 \\
\hline
 &  & $F_{15}$ & $F_{16}$ & $F_{17}$ & $F_{18}$ & $F_{19}$ & $F_{20}$ & -  \\
\hline
 & Best & 1.42e+04 & 4.19e+02 & 2.07e+06 & 8.96e+07 & 8.51e+06 & 1.00e+08 &  \\
 & Median & 1.51e+04 & 4.21e+02 & 2.62e+06 & 2.36e+08 & 1.21e+07 & 2.63e+08 &  \\
FEs = 1.2e5 & Worst & 1.63e+04 & 4.23e+02 & 3.40e+06 & 3.40e+09 & 1.62e+07 & 7.96e+08 &  \\
 & Mean & 1.51e+04 & 4.21e+02 & 2.57e+06 & 3.94e+08 & 1.22e+07 & 2.94e+08 &  \\
 & Std & 4.45e+02 & 9.72e-01 & 2.78e+05 & 6.47e+08 & 1.62e+06 & 1.63e+08 &  \\
\hline
 & Best & 1.27e+04 & 4.15e+02 & 6.87e+05 & 8.79e+03 & 6.73e+06 & 2.96e+03 &  \\
 & Median & 1.36e+04 & 4.19e+02 & 7.87e+05 & 2.15e+04 & 8.95e+06 & 3.54e+03 &  \\
FEs = 6.0e5 & Worst & 1.46e+04 & 4.21e+02 & 9.31e+05 & 3.78e+04 & 1.12e+07 & 1.06e+04 &  \\
 & Mean & 1.36e+04 & 4.19e+02 & 7.96e+05 & 2.14e+04 & 9.09e+06 & 3.88e+03 &  \\
 & Std & 4.31e+02 & 1.07e+00 & 5.84e+04 & 7.52e+03 & 1.10e+06 & 1.47e+03 &  \\
\hline
 & Best & 1.26e+04 & 3.01e+02 & 7.26e+04 & 1.37e+03 & 4.06e+06 & 1.57e+03 &  \\
 & Median & 1.31e+04 & 4.17e+02 & 8.78e+04 & 2.38e+03 & 7.56e+06 & 1.86e+03 &  \\
FEs = 3.0e6 & Worst & 1.38e+04 & 4.19e+02 & 1.11e+05 & 5.57e+03 & 9.02e+06 & 2.44e+03 &  \\
 & Mean & 1.31e+04 & 4.12e+02 & 8.96e+04 & 2.56e+03 & 7.32e+06 & 1.89e+03 &  \\
 & Std & 2.69e+02 & 2.33e+01 & 9.78e+03 & 8.98e+02 & 1.16e+06 & 2.35e+02 &  \\
\hline
\end{tabular}
}
\end{center}
\end{small}
\end{table*}

\begin{table*}[!htb]
\begin{small}
\begin{center}
\caption{Comparison of experimental results at ${FEs}=3.0e+6$}
\label{tab:comp2}
\scalebox{0.7}
{
\begin{tabular}{|c|c|c|c|c|c|c|c|c|}
\hline
 & & $F_{1}$ & $F_{2}$ & $F_{3}$ & $F_{4}$ & $F_{5}$ & $F_{6}$ & $F_{7}$ \\
\hline
 & Best & 1.63e-07 & 1.25e+03 & 1.20e+00 & 7.78e+12 & 1.50e+08 & 3.89e+06 & 4.26e+07 \\
 & Median & 2.86e-07 & 1.31e+03 & 1.39e+00 & 1.51e+13 & 2.38e+08 & 4.80e+06 & 1.07e+08 \\
 DECC-G & Worst & 4.84e-07 & 1.40e+03 & 1.68e+00 & 2.65e+13 & 4.12e+08 & 7.73e+06 & 6.23e+08 \\
 & Mean & 2.93e-07 & 1.31e+03 & 1.39e+00 & 1.70e+13 & 2.63e+08 & 4.96e+06 & 1.63e+08 \\
 & Std & 8.62e-08 & 3.26e+01 & 9.73e-02 & 5.37e+12 & 8.44e+07 & 8.02e+05 & 1.37e+08 \\
\hline
 & Best & 6.33e-12 & 4.21e+02 & 2.23e-08 & 9.76e+11 & 2.08e+08 & 5.07e-03 & 3.45e+06 \\
 & Median & 8.97e-12 & 4.43e+02 & 3.30e-08 & 1.96e+12 & 2.49e+08 & 8.85e-03 & 1.04e+07 \\
 DECC-G$^*$ & Worst & 1.31e-11 & 4.57e+02 & 4.16e-08 & 5.39e+12 & 2.72e+08 & 1.40e-02 & 2.28e+07 \\
 & Mean & 8.81e-12 & 4.42e+02 & 3.30e-08 & 2.29e+12 & 2.45e+08 & 8.77e-03 & 1.10e+07 \\
 & Std & 1.49e-12 & 9.94e+00 & 5.20e-09 & 9.97e+11 & 1.64e+07 & 2.46e-03 & 5.44e+06 \\
\hline
 & Best & 0.00e+00 & 1.73e-11 & 1.28e-13 & 4.27e+12 & 2.15e+08 & 5.85e+06 & 4.16e+04 \\
 & Median & 0.00e+00 & 6.43e-11 & 1.46e-13 & 1.03e+13 & 3.92e+08 & 1.95e+07 & 5.15e+05 \\
 MLCC & Worst & 3.83e-26 & 1.09e+01 & 1.86e-11 & 1.62e+13 & 4.87e+08 & 1.98e+07 & 2.78e+06 \\
 & Mean & 1.53e-27 & 5.57e-01 & 9.88e-13 & 9.61e+12 & 3.84e+08 & 1.62e+07 & 6.89e+05 \\
 & Std & 7.66e-27 & 2.21e+00 & 3.70e-12 & 3.43e+12 & 6.93e+07 & 4.97e+06 & 7.37e+05 \\
\hline
\input{tabPart1a}
\hline
 & & $F_{8}$ & $F_{9}$ & $F_{10}$ & $F_{11}$ & $F_{12}$ & $F_{13}$ & $F_{14}$ \\
\hline
 & Best & 6.37e+06 & 2.66e+08 & 1.03e+04 & 2.06e+01 & 7.78e+04 & 1.78e+03 & 6.96e+08 \\
 & Median & 6.70e+07 & 3.18e+08 & 1.07e+04 & 2.33e+01 & 8.87e+04 & 3.00e+03 & 8.07e+08 \\
 DECC-G & Worst & 9.22e+07 & 3.87e+08 & 1.17e+04 & 2.79e+01 & 1.07e+05 & 1.66e+04 & 9.06e+08 \\
 & Mean & 6.44e+07 & 3.21e+08 & 1.06e+04 & 2.34e+01 & 8.93e+04 & 5.12e+03 & 8.08e+08 \\
 & Std & 2.89e+07 & 3.38e+07 & 2.95e+02 & 1.78e+00 & 6.87e+03 & 3.95e+03 & 6.07e+07 \\
\hline
 & Best & 2.79e+07 & 1.18e+07 & 2.33e+03 & 5.82e-08 & 6.16e+01 & 3.78e+02 & 2.46e+07 \\
 & Median & 4.07e+07 & 1.41e+07 & 2.49e+03 & 7.52e-08 & 7.72e+01 & 5.40e+02 & 2.90e+07 \\
 DECC-G$^*$ & Worst & 1.50e+08 & 1.77e+07 & 2.64e+03 & 8.79e-01 & 1.19e+02 & 7.55e+02 & 3.56e+07 \\
 & Mean & 6.14e+07 & 1.41e+07 & 2.48e+03 & 3.52e-02 & 7.87e+01 & 5.50e+02 & 2.91e+07 \\
 & Std & 3.24e+07 & 1.39e+06 & 7.63e+01 & 1.76e-01 & 1.41e+01 & 9.78e+01 & 2.91e+06 \\
\hline
 & Best & 4.51e+04 & 8.96e+07 & 2.52e+03 & 1.96e+02 & 2.42e+04 & 1.01e+03 & 2.62e+08 \\
 & Median & 4.67e+07 & 1.24e+08 & 3.16e+03 & 1.98e+02 & 3.47e+04 & 1.91e+03 & 3.16e+08 \\
 MLCC & Worst & 9.06e+07 & 1.46e+08 & 5.90e+03 & 1.98e+02 & 4.25e+04 & 3.47e+03 & 3.77e+08 \\
 & Mean & 4.38e+07 & 1.23e+08 & 3.43e+03 & 1.98e+02 & 3.49e+04 & 2.08e+03 & 3.16e+08 \\
 & Std & 3.45e+07 & 1.33e+07 & 8.72e+02 & 6.98e-01 & 4.92e+03 & 7.27e+02 & 2.77e+07 \\
\hline
\input{tabPart2a}
\hline
 & & $F_{15}$ & $F_{16}$ & $F_{17}$ & $F_{18}$ & $F_{19}$ & $F_{20}$ & \\
\hline
 & Best & 1.09e+04 & 5.97e+01 & 2.50e+05 & 5.61e+03 & 1.02e+06 & 3.59e+03 & \\
 & Median & 1.18e+04 & 7.51e+01 & 2.89e+05 & 2.30e+04 & 1.11e+06 & 3.98e+03 & \\
 DECC-G  & Worst & 1.39e+04 & 9.24e+01 & 3.26e+05 & 4.71e+04 & 1.20e+06 & 5.32e+03 & \\
 & Mean & 1.22e+04 & 7.66e+01 & 2.87e+05 & 2.46e+04 & 1.11e+06 & 4.06e+03 & \\
 & Std & 8.97e+02 & 8.14e+00 & 1.98e+04 & 1.05e+04 & 5.15e+04 & 3.66e+02 & \\
\hline
 & Best & 3.62e+03 & 7.04e-08 & 8.09e+01 & 8.37e+02 & 9.90e+05 & 2.83e+03 & \\
 & Median & 3.88e+03 & 1.04e-07 & 1.03e+02 & 1.08e+03 & 1.15e+06 & 3.21e+03 & \\
 DECC-G$^*$ & Worst & 4.25e+03 & 2.18e+00 & 1.33e+02 & 1.53e+03 & 1.23e+06 & 6.23e+03 & \\
 & Mean & 3.88e+03 & 4.01e-01 & 1.03e+02 & 1.08e+03 & 1.14e+06 & 3.33e+03 & \\
 & Std & 1.76e+02 & 6.59e-01 & 1.38e+01 & 1.61e+02 & 5.85e+04 & 6.63e+02 & \\
\hline
 & Best & 5.30e+03 & 2.08e+02 & 1.38e+05 & 2.51e+03 & 1.21e+06 & 1.70e+03 & \\
 & Median & 6.89e+03 & 3.95e+02 & 1.59e+05 & 4.17e+03 & 1.36e+06 & 2.04e+03 & \\
 MLCC & Worst & 1.04e+04 & 3.97e+02 & 1.86e+05 & 1.62e+04 & 1.54e+06 & 2.34e+03 & \\
 & Mean & 7.11e+03 & 3.76e+02 & 1.59e+05 & 7.09e+03 & 1.36e+06 & 2.05e+03 & \\
 & Std & 1.34e+03 & 4.71e+01 & 1.43e+04 & 4.77e+03 & 7.35e+04 & 1.80e+02 & \\
\hline
\input{tabPart3a}
\end{tabular}
}
\end{center}
\end{small}
\end{table*}

Description, characteristics and results of the mentioned algorithms are published on the conference sites~\cite{website:Nical}. The comparison was performed similarly to rules prescribed by the organizing committee of the competition on LSGO CEC'2011. That is, the results for each algorithm are divided into three categories determined by the $\mathit{FEs}$ limits $1.2e5$, $6.0e5$, and $3.0e6$, i.e. at the $\frac{1}{25}$, $\frac{1}{5}$ and the final evaluation numbers. In the absence of results from the first two categories, only the final results have been dealt with. Then, the algorithms in the experiment are ranked according to their decreasing mean-values with ranks from 1-4. Finally, rank 1 is rewarded with 25, rank 2 with 18, rank 3 with 15, and rank 4 with 12 points. The outcome of this estimation is presented in Table~\ref{tab:points}.

\begin{table}[htb]        
\caption{Comparison of algorithms for LSGO}
\label{tab:points}
\vspace{-5mm}
\begin{center}
\begin{tabular}{ l r r r r r r }
\hline
 Alg. & I & II & III & IV & V & Sum. \\
\hline
DECC-G & 39 & 66 & 66 & 66 & 37 & 274 \\
DECC-G* & 51 & 91 & 125 & 125 & 33 & 425 \\
MLCC & 75 & 75 & 75 & 75 & 30 & 330 \\
MABC & 45 & 118 & 84 & 84 & 40 & 371 \\
\hline
\end{tabular}
\end{center}
\vspace{-5mm}
\end{table}

Note that the points in Table~\ref{tab:points} are divided into five groups according their problem classes. The MABC algorithm outperformed the other algorithms by solving the problem classes II and V, whilst DECC-G* the problem classes III and IV, and MLCC the problem class I. In summary, the best results were obtained by DECC-G*. However, the performance of MABC was comparable with DECC-G*.

In order to check how significant these results are, the Friedman nonparametric test as proposed in~\cite{Garcia:2009} was
 performed with significant level $\alpha = 0.05$. According to this test, DECC-G* and MABC significantly improved the results of DECC-G.

\section{Conclusion}
This study presented a MABC algorithm that hybridizes the original ABC using two local search heuristics, i.e. NMA and RWDE. The first is dedicated more to exploration, whilst the second more to exploitation of the search space. An adaptive stochastic function was employed in order to balance the process of exploration/exploitation. The results of MABC were performed in the Special Suite on LSGO in the 2012 IEEE CEC, and compared with the results of algorithms: DECC-G, DECC-G* and MLCC. These results were very competitive when compared with the other algorithms. In addition, this algorithm has been analyzed in the spirit of MC, where four exploration stages are identified: stochastic long-distance, stochastic moderate-distance, deterministic short-distance, and random long-distance explorations. Each of these stages should correspond to the particular meme. However, the interaction between memes leaves additional options for further development of this algorithm. Finally, the adaptive stochastic function for balancing the process of exploration/exploitation bases on phenotypic diversity that could be less important in a well-balanced multi-modal landscapes. In order to show how a genotypic diversity influences the results, other diversity measures should be observed in the future.

\section*{Acknowledgment}

The authors would like to acknowledge the efforts of the organizers of this session and availability of source code of benchmark functions. In addition, we thank to prof. Fatih Tasgetiren for help by searching the new directions in development of the ABC algorithm.

\bigskip{\small \smallskip\noindent Updated 4 June 2012.}
\end{document}

%% file: tabPart1a.tex
 & Best & 4.07e-22 & 5.82e+03 & 1.52e-12 & 4.52e+11 & 8.65e+07 & 7.61e-09 & 2.37e+03 \\
 & Median & 8.86e-22 & 6.08e+03 & 8.95e-01 & 1.44e+12 & 1.09e+08 & 1.98e+00 & 1.04e+04 \\
MABC & Worst & 5.44e-21 & 6.35e+03 & 2.79e+00 & 3.85e+12 & 1.56e+08 & 3.27e+00 & 9.85e+04 \\
 & Mean & 1.63e-21 & 6.09e+03 & 8.20e-01 & 1.64e+12 & 1.13e+08 & 1.86e+00 & 1.36e+04 \\
 & Std & 1.42e-21 & 1.29e+02 & 7.96e-01 & 9.92e+11 & 1.73e+07 & 8.49e-01 & 1.84e+04 \\
\hline

%% file: tabPart2a.tex
 & Best & 1.14e+03 & 3.06e+07 & 1.02e+04 & 1.35e+01 & 1.01e+04 & 5.34e+02 & 1.04e+08 \\
 & Median & 5.91e+05 & 3.74e+07 & 1.05e+04 & 3.17e+01 & 1.27e+04 & 7.99e+02 & 1.23e+08 \\
MABC & Worst & 7.87e+07 & 4.39e+07 & 1.08e+04 & 1.80e+02 & 1.54e+04 & 1.41e+03 & 1.31e+08 \\
 & Mean & 9.77e+06 & 3.76e+07 & 1.04e+04 & 4.25e+01 & 1.28e+04 & 8.53e+02 & 1.22e+08 \\
 & Std & 2.12e+07 & 3.43e+06 & 1.84e+02 & 3.40e+01 & 1.36e+03 & 2.19e+02 & 6.78e+06 \\
\hline

%% file: tabPart3a.tex
 & Best & 1.26e+04 & 3.01e+02 & 7.26e+04 & 1.37e+03 & 4.06e+06 & 1.57e+03 &  \\
 & Median & 1.31e+04 & 4.17e+02 & 8.78e+04 & 2.38e+03 & 7.56e+06 & 1.86e+03 &  \\
MABC & Worst & 1.38e+04 & 4.19e+02 & 1.11e+05 & 5.57e+03 & 9.02e+06 & 2.44e+03 &  \\
 & Mean & 1.31e+04 & 4.12e+02 & 8.96e+04 & 2.56e+03 & 7.32e+06 & 1.89e+03 &  \\
 & Std & 2.69e+02 & 2.33e+01 & 9.78e+03 & 8.98e+02 & 1.16e+06 & 2.35e+02 &  \\
\hline